\documentclass{article}

\usepackage{microtype}
\usepackage{graphicx}
\usepackage{subfigure}
 \usepackage{booktabs}

\usepackage{hyperref}

\usepackage[accepted]{icml2024}

\usepackage{amsmath}
\usepackage{amssymb}
\usepackage{mathtools}
\usepackage{amsthm}

\usepackage[capitalize,noabbrev]{cleveref}

\theoremstyle{plain}

\theoremstyle{definition}

\theoremstyle{remark}

\usepackage[textsize=tiny]{todonotes}

\icmltitlerunning{GPT Semantic Cache: Reducing LLM Costs and Latency via Semantic Embedding Caching}

\begin{document}

\twocolumn[
\icmltitle{ GPT Semantic Cache: Reducing LLM Costs and Latency via Semantic Embedding Caching }




\begin{icmlauthorlist}
\icmlauthor{Sajal Regmi}{}
\icmlauthor{Chetan Phakami Pun}{}
\end{icmlauthorlist}

\icmlcorrespondingauthor{Sajal Regmi}{sajaregmi4@gmail.com}


\vskip 0.3in
]



\printAffiliationsAndNotice{}  

\begin{abstract}
Large Language Models (LLMs), such as GPT \cite{radford2019gpt2}, have revolutionized artificial intelligence by enabling nuanced understanding and generation of human-like text across a wide range of applications. However, the high computational and financial costs associated with frequent API calls to these models present a substantial bottleneck, especially for applications like customer service chatbots that handle repetitive queries. In this paper, we introduce GPT Semantic Cache, a method that leverages semantic caching of query embeddings in in-memory storage (Redis). By storing embeddings of user queries, our approach efficiently identifies semantically similar questions, allowing for the retrieval of pre-generated responses without redundant API calls to the LLM.This technique achieves a notable reduction in operational costs while significantly enhancing response times, making it a robust solution for optimizing LLM-powered applications. Our experiments demonstrate that GPT Semantic Cache reduces API calls by up to \textbf{68.8\%} across various query categories, with cache hit rates ranging from \textbf{61.6\%} to \textbf{68.8\%}. Additionally, the system achieves high accuracy, with positive hit rates exceeding \textbf{97\%}, confirming the reliability of cached responses. This technique not only reduces operational costs, but also improves response times, enhancing the efficiency of LLM-powered applications.
\end{abstract}

\section{Introduction}
\label{submission}
Large Language Models (LLMs) such as GPT have become integral to modern AI applications due to their ability to understand and generate human-like text. They are widely used in chatbots, virtual assistants, and customer support systems to interpret user queries and provide relevant responses.

Despite their capabilities, a significant challenge arises from the need to make individual API calls to the LLM for each user query. This process can be both time-consuming and costly, particularly when dealing with large volumes of similar or repetitive questions common in customer service scenarios. For instance, customer support chatbots frequently encounter repetitive queries with minor variations, leading to redundant processing and elevated computational expenses.

To address this inefficiency, various methods have been proposed to cache LLM responses. For instance, \cite{bang-2023-gptcache} introduced {GPTCache, a Python package designed to cache LLM responses to reduce latency and costs. Similarly, \cite{chen2023acceleratinglargelanguagemodel} explored techniques to improve the efficiency of LLM decoding employing speculative sampling (SpS), an approach that accelerates token generation using a draft model and rejection sampling. While their work focuses on reducing latency during sequence generation, our GPT Semantic Cache complements these efforts by addressing latency and cost issues through semantic caching of query embeddings, enabling efficient retrieval of responses for semantically similar queries.

Building upon these ideas, we propose GPT Semantic Cache. By converting user queries into numerical embedding representations that capture the semantic meaning, we can store and quickly retrieve these embeddings \cite{openai2022textembeddings}. When a new query arrives, we compare its embedding to those stored in the cache. If a semantically similar query is found, we can provide the corresponding response immediately, bypassing the need for an additional API call to the LLM.

This approach offers several advantages::
\begin{itemize}
\item \textbf{Reduced Latency}: Users receive faster responses since the system can retrieve the responses directly from the cache.

\item \textbf{Cost Efficiency}: Fewer API calls to the LLM result in lower operational costs.

\item \textbf{Scalability}: The system can handle higher volumes of queries without a proportional increase in computational resources.

\end{itemize}

\section{System Architecture}

The architecture of the \textbf{GPT Semantic Cache} system is designed to optimize the handling of repeated and semantically similar queries, reducing costs and improving response times. The system comprises several key components, each playing a crucial role in the overall workflow.

\subsection{Overview}

The architecture consists of three main components:

\begin{itemize} \item \textbf{Embedding Generation:} Converts user queries into semantic embeddings. \item \textbf{In-Memory Caching:} Manages storage and retrieval of embeddings and responses using Redis. \item \textbf{Similarity Search:} Identifies semantically similar queries using Approximate Nearest Neighbor (ANN) techniques. \end{itemize}

\begin{figure}[ht]
\vskip 0.2in
\begin{center}
\centerline{\includegraphics[width=\columnwidth]{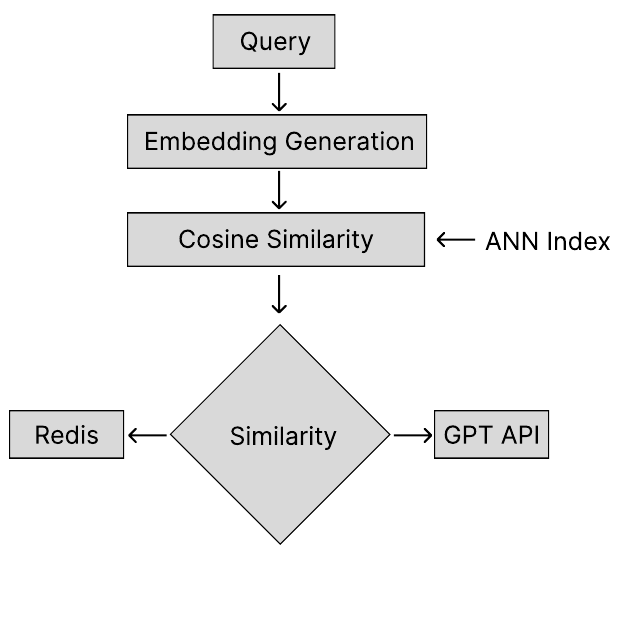}}
\caption{The diagram illustrates the core components of the GPT Semantic Cache system, showcasing the flow of user queries through embedding generation, similarity calculation \cite{rahutomo2012semantic}, and Approximate Nearest Neighbors (ANN) indexing. The Redis-based in-memory cache stores embeddings and corresponding responses, facilitating quick retrieval. Queries are sent to the GPT API only when no matching response is found in the cache.}
\label{icml-historical}
\end{center}
\vskip -0.2in
\end{figure}

\subsection{Embedding Generation}

The system can utilize either the OpenAI API for text embeddings or any ONNX-compatible model from platforms like Hugging Face. This flexibility allows the system to adapt to different deployment scenarios, whether using cloud-based models or self-hosted options. In the embedding generation process:

\begin{itemize} \item \textbf{OpenAI API:} When using OpenAI embeddings, the text is sent to the API, and the response includes the embedding vector. The vector's length is typically 1536 for models like \texttt{text-embedding-ada-002} \cite{openai2022textembeddings}. \item \textbf{Local Models:} For local embeddings, the system loads ONNX-supported models and generates embeddings using libraries such as \texttt{@xenova/transformers}. The generated embeddings are normalized and pooled to create a consistent representation. \end{itemize}

\subsection{In-Memory Caching}

The caching layer uses Redis \cite{redis2022} to store embeddings and their corresponding responses. Redis provides a high-performance, in-memory data store, ensuring quick access and efficient management of cached data. The cache is organized by:

\begin{itemize} \item \textbf{Embedding Size:} Different embedding models may produce vectors of varying dimensions, so the cache is partitioned based on the embedding size. \item \textbf{Time-To-Live (TTL):} Cached data is assigned a TTL to prevent outdated information from persisting indefinitely. This feature ensures that the cache remains fresh and relevant, automatically expiring old entries. \end{itemize}

\subsection{Similarity Search using ANN}

The system uses Approximate Nearest Neighbor (ANN) algorithms, such as Hierarchical Navigable Small World (HNSW) graphs \cite{malkov2018efficient} implemented via \texttt{hnswlib-node}, to perform efficient similarity searches. HNSW constructs a multi-layer graph $G = (V, E)$, where nodes $V$ correspond to stored embeddings, and edges $E$ represent proximity connections. The process involves:

\begin{itemize} \item \textbf{Index Initialization:} When the cache is loaded, the system initializes the ANN index with stored embeddings. If the index has not been initialized, it starts with a minimal size and dynamically grows as new data is added.

 For an embedding $\mathbf{e}$, its neighbors in the graph are chosen based on the top-$k$ similarity scores:
    \[
    \text{neighbors}(\mathbf{e}) = \text{arg top-}k \big( \text{cosine\_similarity}(\mathbf{e}, \mathbf{e}_i) \big)
    \]

\item \textbf{Embedding Search:} When a new query embedding is generated, the ANN index searches for the $k$ most similar embeddings using cosine similarity. This approach ensures rapid and scalable similarity matching. For a new query embedding $\mathbf{q}$, the search involves a multi-layer traversal of the HNSW graph. Starting from an entry point, the algorithm iteratively explores closer neighbors based on cosine similarity until it converges to the top-$k$ closest nodes:
    \[
    \text{result\_set} = \text{arg top-}k \big( \text{cosine\_similarity}(\mathbf{q}, \mathbf{e}_i) \big)
    \]

Compared to exhaustive search, HNSW reduces complexity from $O(n)$ to approximately $O(\log n)$ for search queries, where $n$ is the number of embeddings. This efficiency is achieved through hierarchical partitioning and greedy search strategies \cite{malkov2018efficientrobustapproximatenearest}.

To ensure robustness in handling dynamic updates, our system periodically rebalances the HNSW graph, maintaining search efficiency even as new embeddings are added. Future iterations will explore adaptive indexing strategies for further performance gains.

\end{itemize}

\subsection{Query Handling Workflow}

\begin{enumerate} \item \textbf{Cache Lookup:} When a query is received, it is first converted into an embedding. The system then searches the ANN index for similar embeddings. If a match with a similarity score above a specified threshold (e.g. 0.8) is found, the cached response is returned. \item \textbf{Cache Miss:} If no suitable match is found, the system sends the query to the LLM API (e.g., OpenAI GPT) to generate a new response. The query, embedding, and response are then stored in the cache for future use. \item \textbf{Embedding Storage:} New embeddings and their responses are added to both the ANN index and the Redis cache. The system manages the cache size and ensures that the index can accommodate growing data by resizing when necessary. \end{enumerate}

\subsection{Cosine Similarity and Similarity Threshold}

Cosine similarity is employed to measure the similarity between query embeddings.

Given two embeddings $\mathbf{u}, \mathbf{v} \in \mathbb{R}^d$, the cosine similarity between them is defined as:

\[
\text{cosine\_similarity}(\mathbf{u}, \mathbf{v}) = \frac{\mathbf{u} \cdot \mathbf{v}}{\|\mathbf{u}\| \|\mathbf{v}\|},
\]

where $\mathbf{u} \cdot \mathbf{v}$ is the dot product, and $\|\mathbf{u}\|$ and $\|\mathbf{v}\|$ are the Euclidean norms of $\mathbf{u}$ and $\mathbf{v}$, respectively.

In the caching system, cosine similarity is used to determine the proximity between a query embedding $\mathbf{q}$ and stored embeddings $\{\mathbf{e}_1, \mathbf{e}_2, \ldots, \mathbf{e}_n\}$, enabling the retrieval of the most relevant cached response. 

The calculated cosine of the angle between two vectors, provides a value between $-1$ and $1$, where $1$ indicates identical orientation. In the context of semantic embeddings, a higher similarity score implies greater semantic closeness. The system uses a similarity threshold (0.8) to decide whether a cached response is sufficiently similar to the new query to be reused.

\subsection{Cache Management with Time-To-Live (TTL)}

To prevent the cache from becoming stale, each cached entry is assigned a Time-To-Live (TTL) value. The TTL ensures that data is automatically removed after a certain period, maintaining the relevance of cached responses. This mechanism also helps manage the cache size, preventing it from growing indefinitely.

\subsection{System Workflow Summary}

The overall workflow of the GPT Semantic Cache system is as follows:

\begin{enumerate} \item \textbf{Receive Query:} The user submits a query to the system. \item \textbf{Generate Embedding:} The query is converted into an embedding using either the OpenAI API or a local ONNX model. \item \textbf{Search Cache:} The embedding is used to search the ANN index for similar embeddings. \item \textbf{Determine Similarity:} Cosine similarity is calculated between the query embedding and cached embeddings. \item \textbf{Retrieve or Generate Response:} \begin{itemize} \item \textbf{Cache Hit:} If a similar embedding is found above the threshold, the corresponding cached response is returned. \item \textbf{Cache Miss:} If no similar embedding is found, the query is sent to the LLM API to generate a new response, which is then cached. \end{itemize} \item \textbf{Update Cache and ANN Index:} The new embedding and response are stored in Redis and added to the ANN index for future queries. \end{enumerate}

\subsection{Advantages of the Architecture}

The GPT Semantic Cache system offers several benefits:

\begin{itemize} \item \textbf{Reduced Latency:} By serving responses from the cache when possible, the system provides faster response times to users. \item \textbf{Cost Savings:} Reducing the number of API calls to the LLM lowers operational costs significantly. \item \textbf{Scalability:} The use of ANN and in-memory caching allows the system to handle increasing volumes of queries efficiently. \item \textbf{Flexibility:} Support for multiple embedding models and configurations makes the system adaptable to various deployment needs. \end{itemize}

\subsection{Potential Extensions}

Future enhancements to the system could include:

\begin{itemize} \item \textbf{Dynamic Threshold Adjustment:} Implementing mechanisms to adjust the similarity threshold based on system performance or user feedback. \item \textbf{Distributed Caching:} Expanding the caching layer to a distributed setup for higher availability and fault tolerance. \item \textbf{Advanced Embedding Models:} Incorporating newer or domain-specific embedding models to improve semantic understanding. \end{itemize}

\section{Testing Methodology}

To evaluate the effectiveness of our GPT Semantic Cache system, we conducted a series of experiments designed to simulate real-world user interactions with AI-based services. The primary objectives were to assess the system's ability to efficiently handle repeated or semantically similar queries, validate the accuracy of cached responses, and quantify improvements in both response time and operational cost.

\subsection{Dataset Preparation and Cache Population}

We constructed a dataset of 8,000 question-answer pairs across four categories: basic Python programming, customer service inquiries, technical support questions, and general knowledge. These categories were selected to reflect the diverse range of queries typically encountered in real-world applications. Each question was converted into a semantic embedding using all-MiniLM-L6-v2
\cite{reimers2020allminilm}, and both the embeddings and their corresponding responses were stored in Redis. An Approximate Nearest Neighbor (ANN) index was also updated to include these embeddings for efficient similarity searches.

\subsection{Test Query Generation and Execution}
To evaluate the system's performance, we generated an additional 2,000 test queries, with 500 queries per category. Each query was submitted to the system, where it was converted into an embedding and compared against cached embeddings using the ANN index. If a similar embedding was found (similarity $>=$ 0.8), the corresponding cached response was returned (cache hit). Otherwise, the query was forwarded to the LLM for a new response (cache miss), which was subsequently cached.

\subsection{Validation of Cache Hits}

To verify the appropriateness of cached responses, we employed GPT-4o Mini to evaluate whether the retrieved cached response was valid for the test query. For each cache hit, both the test query and the original cached question were provided to the model, which returned a binary verdict indicating whether the queries were semantically similar and whether the cached response was accurate. This validation step enabled us to quantify the accuracy of the caching mechanism. Additionally, the response times for queries were measured and averaged for both the caching system and the traditional method (without cache). This allowed us to determine the performance improvement in terms of query response time.

\section{Results}
The evaluation of GPT Semantic Cache highlights its significant impact on reducing API calls and improving response times. This section provides a detailed analysis of system performance, comparing the traditional method to our caching approach.

\subsection{Reduction in API Calls}

GPT Semantic Cache reduces the number of API calls by serving responses from the cache for semantically similar queries. As shown in Figure 1, the traditional method results in 100\% API calls across all categories, whereas GPT Semantic Cache reduces API dependency significantly. The reduction rates for each category are as follows:

\begin{itemize} 
\item \textbf{Basics of Python Programming:} 67\% cache hit rate, reducing API calls to 33\%.
\item \textbf{Technical Support Related to Network:} 67\% cache hit rate, reducing API calls to 33\%. \item \textbf{Questions Related to Order and Shipping:} 68.8\% cache hit rate, reducing API calls to 31.2\%. \item \textbf{Customer Shopping QA:} 61.6\% cache hit rate, reducing API calls to 38.4\%. 
\end{itemize}

\begin{figure}[ht]
\vskip 0.2in
\begin{center}
\centerline{\includegraphics[width=\columnwidth]{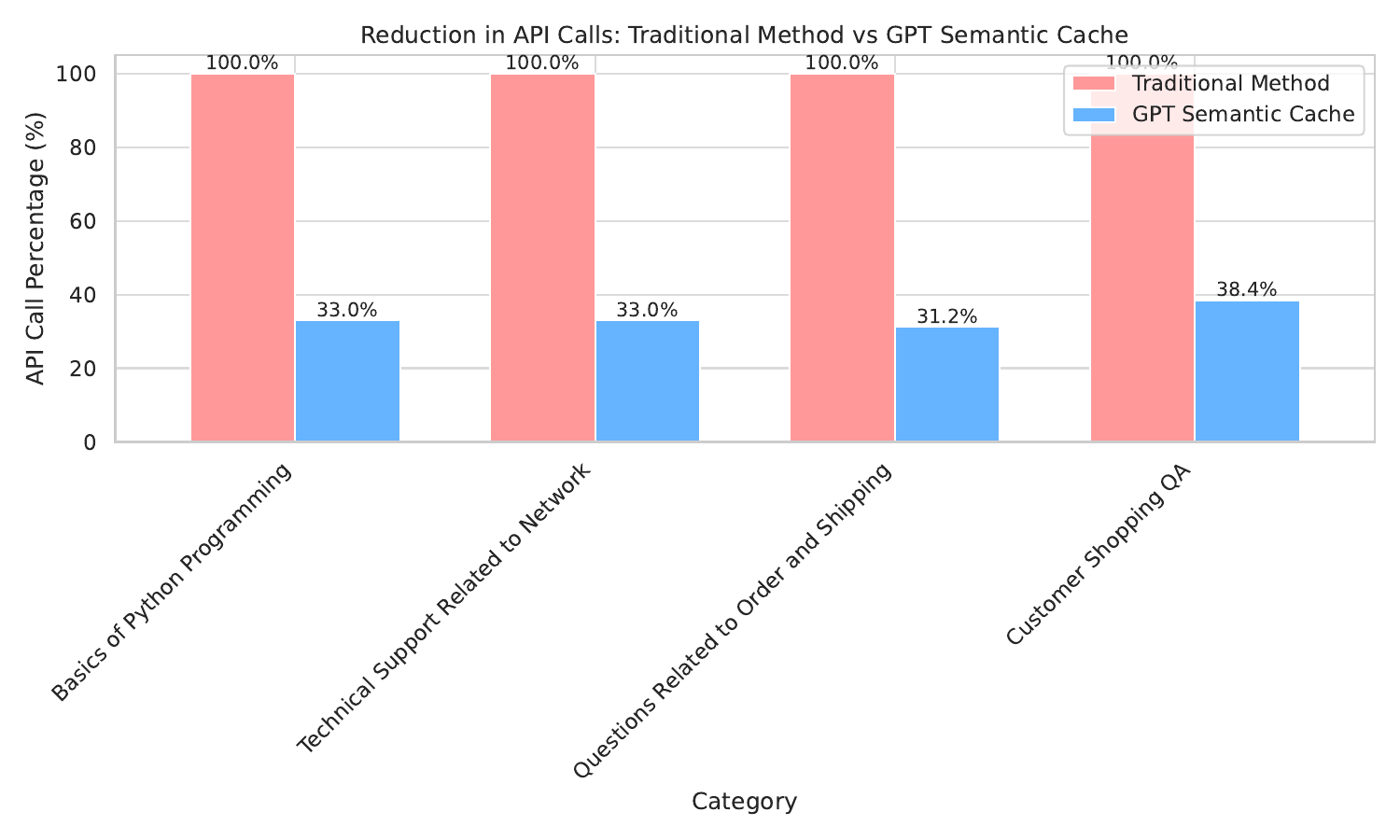}}
\caption{Comparison of API Call Frequency: Traditional Query Handling vs Semantic Caching System.}
\label{icml-historical}
\end{center}
\vskip -0.2in
\end{figure}

\subsection{Response Time Analysis}

To assess performance improvements, we measured and compared the average query response times for both the caching system and the traditional method. Figure 3 illustrates the substantial reduction in response times achieved through GPT Semantic Cache.

\begin{figure}[ht]
\vskip 0.2in
\begin{center}
\centerline{\includegraphics[width=\columnwidth]{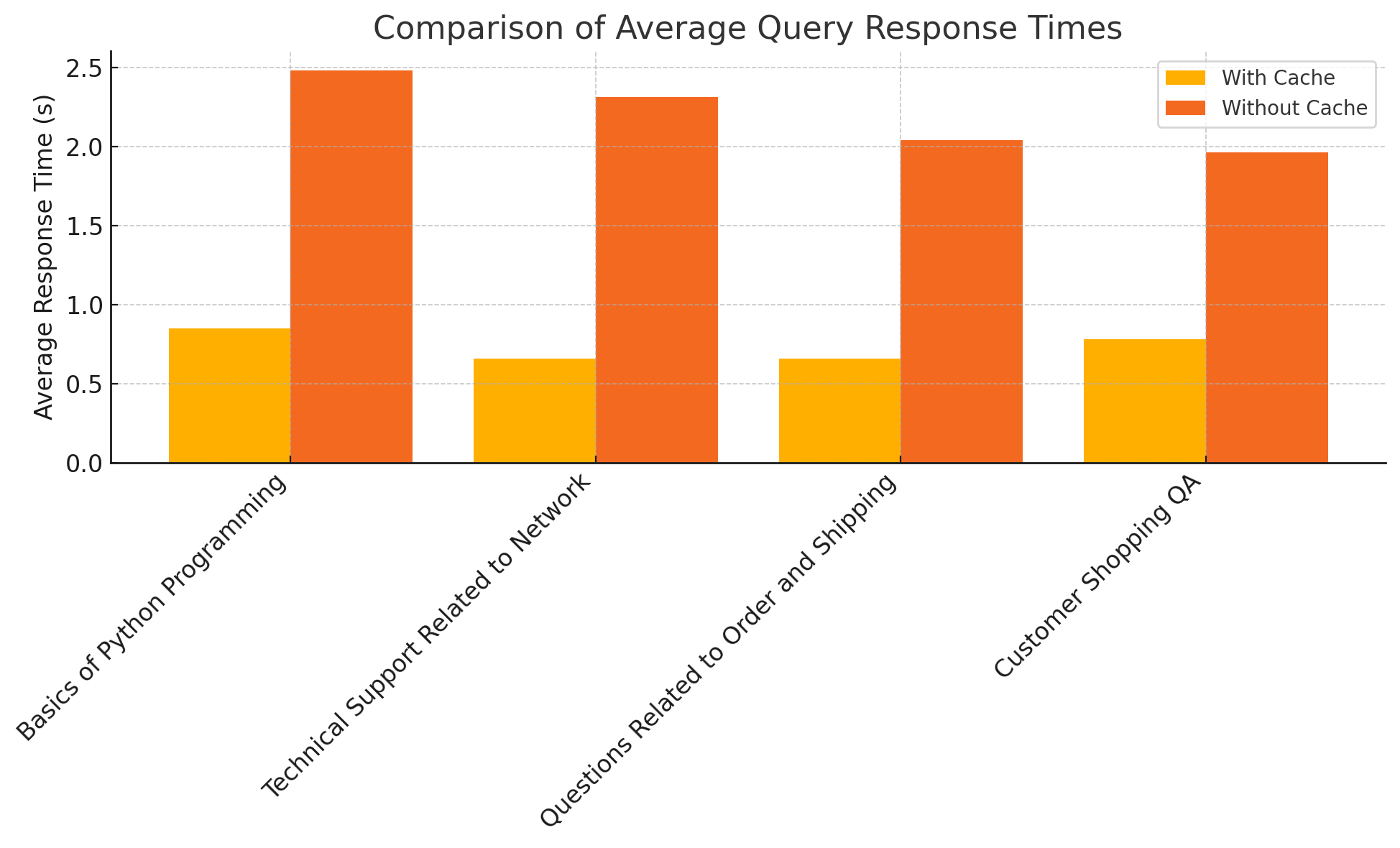}}
\caption{Comparison of Average Query Response Times: With Cache vs Without Cache.}
\label{response-time-comparison}
\end{center}
\vskip -0.2in
\end{figure}

\subsection{Cache Hit and Accuracy Analysis}

The cache hit rates indicate the system's efficiency in recognizing semantically similar queries, while the positive hit rates highlight the accuracy of responses retrieved from the cache. As presented in Figure 2, the system achieved high positive hit rates across all categories, ranging from 92.5\% to 97.3\%, confirming the reliability of cached responses.

\begin{figure}[ht]
\vskip 0.2in
\begin{center}
\centerline{\includegraphics[width=\columnwidth]{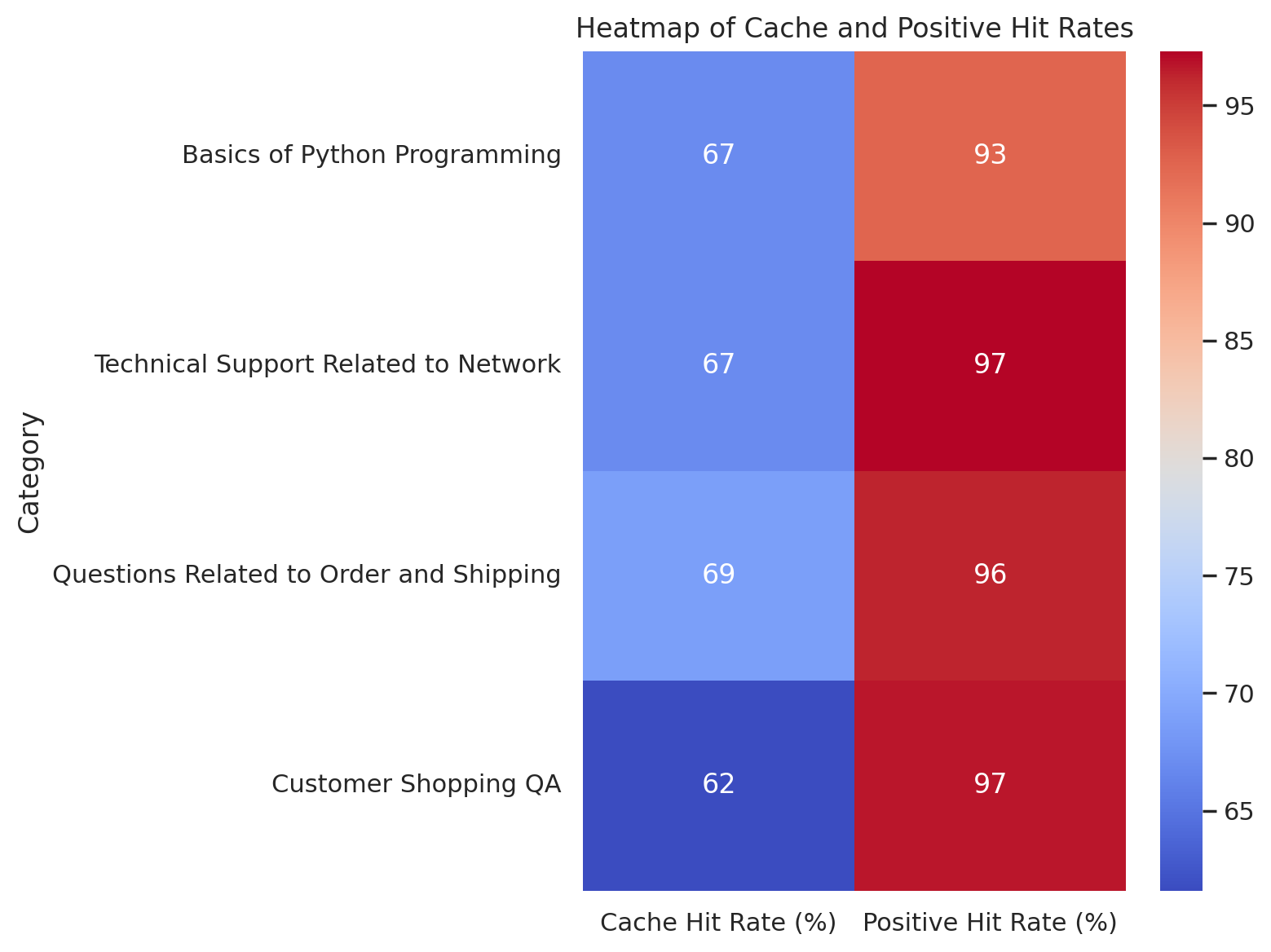}}

\caption{Cache Hit Rates and Positive Match Accuracy Across Query Categories.}
\label{icml-historical}
\end{center}
\vskip -0.2in
\end{figure}

\begin{table}[ht]
\caption{ Cache Hit per 500 queries in each category and the no of positive hits }
\label{sample-table}
\vskip 0.15in
\begin{center}
\begin{small}
\begin{sc}
\begin{tabular}{p{3cm}p{2cm}p{2cm}}
\toprule
Category & Cache Hit & Positive Hits \\
\midrule
Basics of Python Programming & 335 & 310 \\
Technical Support Related to Network & 335 & 326 \\
Questions Related to Order and Shipping & 344 & 331 \\
Customer Shopping QA & 308 & 298 \\

\bottomrule
\end{tabular}
\end{sc}
\end{small}
\end{center}
\vskip -0.1in
\end{table}

\vskip -0.1in

\section{Discussion}

The results of the GPT Semantic Cache demonstrate a significant reduction in API calls and improvement in response times across all evaluated query categories. The cache hit rates of up to 68.8\% and positive hit rates of the cache hit rates exceeding 97\% highlight the system's effectiveness in identifying and leveraging semantic similarity for query handling. However, it is essential to contextualize these outcomes within both their strengths and potential limitations.

 \subsection{Strengths and Contributions}
  The high cache hit and positive hit rates confirm the robustness of the semantic embedding approach combined with an in-memory caching system. Specifically, the system shows remarkable efficiency in categories like "Order and Shipping Queries," where user questions tend to follow predictable patterns, leading to higher semantic overlaps. This efficiency validates the underlying design, making the system highly suitable for applications like customer support chatbots, where such patterns are prevalent.

Additionally, the operational cost savings achieved by reducing LLM API calls are substantial. The use of time-to-live (TTL) mechanisms ensures that cached data remains relevant, addressing a common concern of response accuracy over time. Moreover, by supporting both cloud-based (OpenAI API) and local ONNX models for embedding generation, the system demonstrates flexibility and adaptability for diverse deployment scenarios.

\subsection{Challenges and Limitations}
While the system performed well in categories with structured or repetitive queries, certain limitations were observed in more diverse query types, such as "Customer Shopping QA." This category exhibited slightly lower cache hit rates (61.6\%), likely due to the broader variety of customer inquiries and the greater semantic variability in the language used. In such cases, the fixed similarity threshold (0.8) may exclude some valid matches. Adaptive thresholding, tailored to specific categories or use cases, could mitigate this issue and improve hit rates further.

Maintaining the ANN index for large-scale datasets presents a computational overhead, particularly when embeddings grow into millions of entries. Although the current implementation with HNSW is efficient, scenarios with frequent updates or dynamic embeddings (e.g., real-time user input streams) might introduce latency or consistency challenges. A hybrid caching approach, combining semantic caching with heuristic-based query grouping, could enhance 

\subsection{Threshold Selection (0.8 as Optimal Value)}

The threshold value of 0.8 for cosine similarity was determined empirically through extensive testing on a validation dataset. The experiments involved varying the threshold from 0.6 to 0.9 in increments of 0.05 and measuring the impact on cache hit rates and response accuracy. The results revealed that:

\begin{itemize}
    \item Thresholds below 0.8 led to higher cache hit rates but also introduced irrelevant matches, decreasing the positive hit rate and, consequently, the overall response accuracy.
    \item Thresholds above 0.8 reduced cache hit rates significantly, as fewer queries were identified as semantically similar, leading to more LLM API calls and higher costs.
    \item At 0.8, the system achieved an optimal balance, maintaining a high cache hit rate (up to 68.8\%) without compromising response accuracy (positive hit rates exceeding 97\%).
\end{itemize}

This threshold was thus selected as the optimal trade-off between efficiency and accuracy, aligning with the system’s objectives.
performance in such environments.

\subsection{Scalability and Future Enhancements}
The architecture demonstrates strong scalability in its current form, with Redis providing fast in-memory storage and HNSW ensuring efficient similarity searches. However, deploying the system across distributed environments with multiple nodes may require additional optimizations to manage data consistency and minimize synchronization delays. Integrating advanced embedding models, particularly domain-specific ones, could further enhance semantic understanding and accuracy in specialized applications.

Additionally, dynamic adjustment of similarity thresholds based on query patterns, user feedback, or system performance metrics could refine cache hit rates over time. Incorporating mechanisms to detect and handle edge cases, such as highly ambiguous queries or those requiring context beyond semantic similarity, would further enhance the system’s reliability.

\section{Use Cases}
GPT Semantic Cache is highly applicable in domains where repetitive or similar queries are common. Key use cases include:

\subsection{Customer Support Chatbots}
Frequently asked questions generate redundant queries to LLM APIs, increasing costs and response times. By caching embeddings of past queries and responses, the system retrieves answers directly for semantically similar questions, bypassing LLM calls. For example, a banking chatbot can instantly answer repeated queries like ``What are the interest rates for savings accounts?'' or ``How do I reset my online banking password?'' from the cache.

\subsection{Retrieval-Augmented Generation (RAG) Systems}
RAG systems frequently query LLMs for context-specific answers, leading to high latency. GPT Semantic Cache allows these systems to reuse cached results for semantically similar queries, reducing API calls and accelerating responses. For instance, a document search system can avoid redundant LLM invocations for similar summaries like ``Summarize the financial trends for Q3 2024.''

\subsection{Real-Time Code Assistants}
Developers often generate similar queries for code completion or debugging. By caching embeddings of code-related queries and solutions, the system enables rapid reuse for identical or similar prompts. A query like ``How do I write a function to reverse a string in Python?'' can reuse cached responses for prompts like ``Python function to reverse text.''

\subsection{E-Commerce Product Support}
E-commerce platforms receive redundant queries about product features and availability. Embedding-based caching enables quick retrieval of responses to semantically similar questions, improving customer support efficiency. Questions like ``What are the features of this phone?'' and ``Does this phone have a dual camera?'' can utilize cached responses.

\section{Conclusion}
This paper introduced GPT Semantic Cache, a system that leverages semantic embeddings to cache and retrieve responses from Large Language Models (LLMs) for semantically similar queries. By converting user queries into embeddings and storing them with corresponding responses in an in-memory cache, the system efficiently identifies similar questions and provides immediate answers without redundant API calls to the LLM.

Our experiments demonstrated significant reductions in API calls—up to 68.8\% across various query categories—with cache hit rates ranging from 61.6\% to 68.8\%.The system demonstrated exceptional accuracy, with positive hit rates exceeding 97\%, indicating that the cached responses closely aligned with user intent in most cases. These results highlight the system's effectiveness in reducing operational costs and improving response times, making it highly beneficial for LLM-powered applications dealing with repetitive queries.

GPT Semantic Cache offers a scalable and cost-effective solution for enhancing the efficiency of AI applications such as customer support chatbots, RAG-based systems, real-time code assistants, and e-commerce platforms. Its flexibility in supporting multiple embedding models and configurations allows for adaptation across various domains.

Future work will explore dynamic threshold adjustments for similarity matching, distributed caching mechanisms for improved scalability, and the integration of domain-specific embedding models to enhance semantic understanding in specialized applications.

\section{Limitations}
While GPT Semantic Cache significantly reduces LLM costs and latency, it has limitations. The fixed similarity threshold (0.8) may not generalize across all use cases, potentially missing nuanced query variations. Scalability to massive datasets could degrade as the embedding set grows, introducing latency in dynamic updates. Domain-specific queries may require tailored embeddings for improved accuracy. Additionally, the reliance on TTL for cache freshness is insufficient for rapidly changing data, and the system does not handle context-dependent multi-turn interactions. Resource overhead for cache management might also pose challenges in constrained environments.

\clearpage


\bibliography{GPT_Semantic_Cache}
\bibliographystyle{icml2024}

\newpage
\appendix
\onecolumn
\section{Appendix}

The Package, Dataset and Code used to write this paper

\begin{itemize}
    \item \textbf{Package}: \href{https://www.npmjs.com/package/gpt-semantic-cache}{Test Script: https://www.npmjs.com/package/gpt-semantic-cache}
    \item \textbf{Dataset}: \href{https://github.com/sajalregmi/gpt-semantic-cache-test/tree/main/test_dataset}{Test Dataset: https://github.com/sajalregmi/gpt-semantic-cache-test/tree/main/test\_dataset}
    \item \textbf{Test Code}: \href{https://github.com/sajalregmi/gpt-semantic-cache-test/blob/main/test.ts}{Test Script: https://github.com/sajalregmi/gpt-semantic-cache-test/blob/main/test.ts}

\end{itemize}
\end{document}